\pdfoutput=1

\documentclass[11pt]{article}

\usepackage{acl}

\usepackage{times}
\usepackage{latexsym}

\usepackage[T1]{fontenc}

\usepackage[utf8]{inputenc}

\usepackage{microtype}

\usepackage{inconsolata}

\usepackage{graphicx}

\usepackage{multirow}
\usepackage{booktabs}
\usepackage{amsmath}
\usepackage{amsfonts}
\usepackage{textcomp}

\title{SDPO: Segment-Level Direct Preference Optimization for Social Agents}

\author{\quad Aobo Kong$^{1}$\thanks{~Equal contribution. Work done during internship.} \quad Wentao Ma$^{2}\footnotemark[1]$ \quad Shiwan Zhao$^{1}$ \quad Yongbin Li$^{2}$\thanks{~Corresponding authors.} \quad Yuchuan Wu$^{2}$\\
\textbf{\quad Ke Wang$^{2}$ \quad Xiaoqian Liu$^{2}$ \quad Qicheng Li$^{1}$\footnotemark[2] \quad Yong Qin$^{1}$ \quad Fei Huang$^{2}$}\\
$^1$TMCC, CS, Nankai University \quad $^2$Tongyi Lab\\
\texttt{$^{1}$kongaobo@mail.nankai.edu.cn \quad $^{1}$zhaosw@gmail.com} \\
\texttt{$^{1}$\{liqicheng, qinyong\}@nankai.edu.cn}\\
\texttt{$^{2}$\{mawentao.mwt, shengxiu.wyc, shuide.lyb\}@alibaba-inc.com}\\
}

\begin{document}
\maketitle
\begin{abstract}
Social agents powered by large language models (LLMs) can simulate human social behaviors but fall short in handling complex social dialogues. Direct Preference Optimization (DPO) has proven effective in aligning LLM behavior with human preferences across various agent tasks. However, standard DPO focuses solely on individual turns, which limits its effectiveness in multi-turn social interactions. 
Several DPO-based multi-turn alignment methods with session-level data have shown potential in addressing this problem.
While these methods consider multiple turns across entire sessions, they are often overly coarse-grained, introducing training noise, and lack robust theoretical support. To resolve these limitations, we propose Segment-Level Direct Preference Optimization (SDPO), which dynamically select key segments within interactions to optimize multi-turn agent behavior. 
SDPO minimizes training noise and is grounded in a rigorous theoretical framework. 
Evaluations on the SOTOPIA benchmark demonstrate that SDPO-tuned agents consistently outperform both existing DPO-based methods and proprietary LLMs like GPT-4o, underscoring SDPO's potential to advance the social intelligence of LLM-based agents. We release our code and data at this \href{https://github.com/AlibabaResearch/DAMO-ConvAI/tree/main/SDPO}{url}. 
\end{abstract}

\section{Introduction}

Recent advancements in large language models (LLMs) have significantly enhanced their capabilities in language understanding and generation, particularly within the realm of human-machine interaction. By incorporating identity-specific information, LLM-based agents can simulate human social behaviors, demonstrating basic social intelligence in tasks such as role-playing casual conversations \cite{wang-etal-2024-rolellm, lu-etal-2024-large} and navigate simulated social environments \cite{park2023generative}. However, recent studies \cite{zhou2024sotopia} have shown that, in more complex, goal-oriented social scenarios, such as negotiation, competition, and cooperation, LLMs still struggle to exhibit the nuanced decision-making abilities that are characteristic of human social interactions.

In response to these challenges, several methods have been developed to better align LLM behavior with human preferences. These approaches offer promising strategies for improving social decision-making in LLMs. Specifically, we focus on Direct Preference Optimization (DPO)-based methods. Standard DPO \cite{rafailov2023direct}, identifies a single conversational turn and uses a `positive-negative' pair of responses from that turn to optimize the model via a preference loss function. While DPO has demonstrated some effectiveness, its focus on individual turns limits its ability to model goal completion in goal-oriented social dialogues, where success often relies on high-quality interactions spanning multiple conversational turns.

To more effectively align agent behavior in multi-turn interactions, several multi-turn alignment methods including ETO \cite{song-etal-2024-trial} and DMPO \cite{shi-etal-2024-direct} have been proposed. These methods extend the sampling scope from individual turns to entire sessions, constructing pairs of good and bad sessions and applying an adapted DPO loss for training. We categorize these methods as session-level DPO. However, they exhibit limitations in both data granularity and theoretical foundations. From the data perspective, session-level DPOs suffer from the following drawbacks due to their relatively coarse alignment granularity:

(\romannumeral1) Turns without errors in negative sessions are also treated as bad outputs, introducing substantial noise that negatively affects the training process.

(\romannumeral2) Sampling from scratch provides the interlocutor with a vast action space. A higher score for a positive session may result from changes in the interlocutor's behavior, making it challenging for the model to learn the correct behavior pattern from the positive sample.

From a theoretical perspective, in multi-turn scenarios, simply applying DPO fails to eliminate the partition function $Z$ \cite{shi-etal-2024-direct}. ETO extends the DPO loss to multi-turn interactions but lacks a formal theoretical guarantee. DMPO incorporates SAOM theory and successfully converts $Z$ into a constant; however, due to the differing number of turns between positive and negative sessions, DMPO heuristically normalizes the length to eliminate $Z$ without a rigorous proof. A detailed theoretical analysis is provided in Appendix  \ref{sec: app_dmpo}.

\begin{figure}[t]
\centering
\scalebox{1.65}{
\includegraphics[width=1\textwidth,trim=0.1cm 3.8cm 3.4cm 0cm, clip]{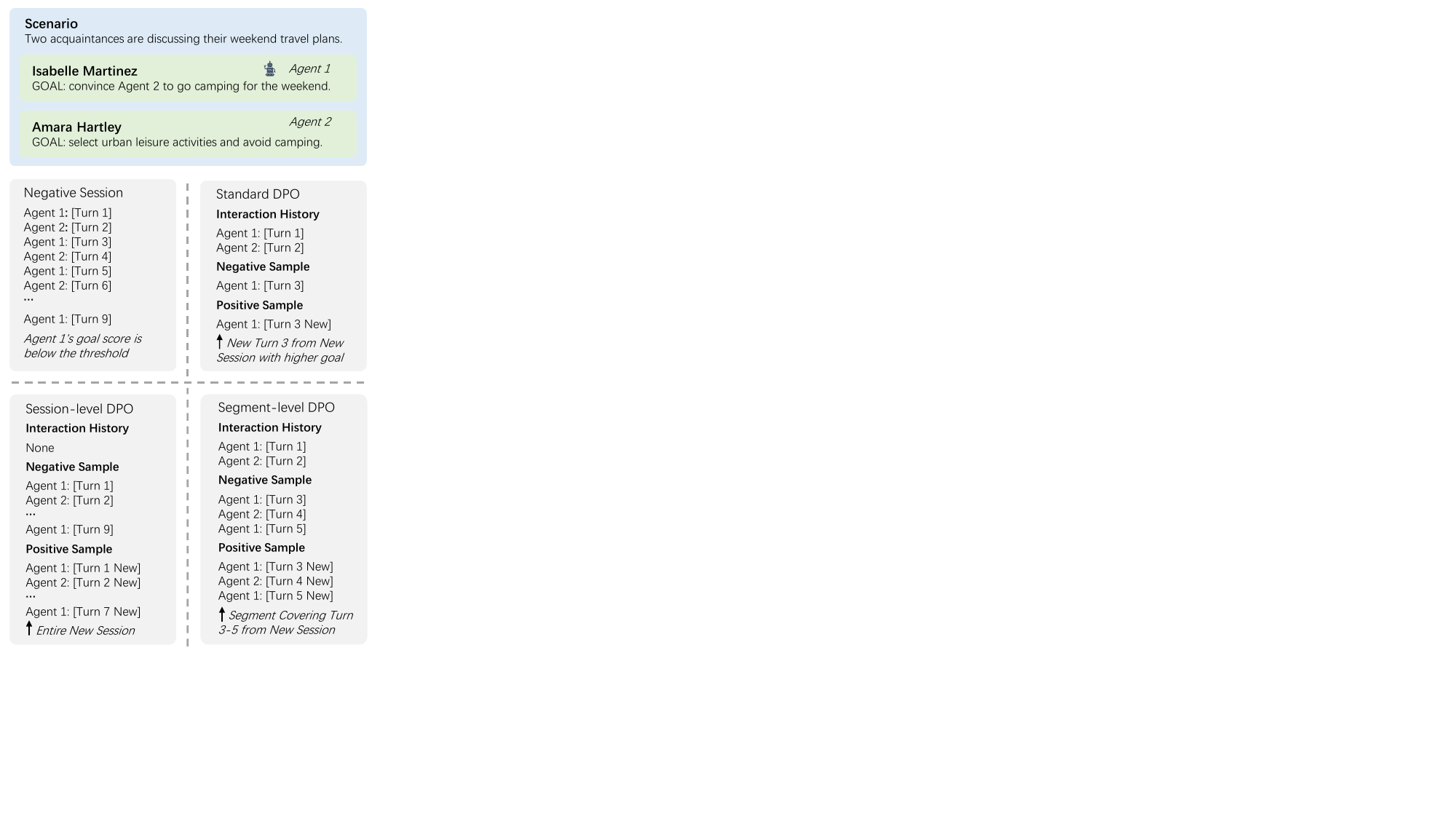}}
\caption{An overview of three alignment algorithms, illustrated using a SOTOPIA task as an example. \includegraphics[scale=0.03, trim=1cm 5.5cm 20cm 2cm, clip]{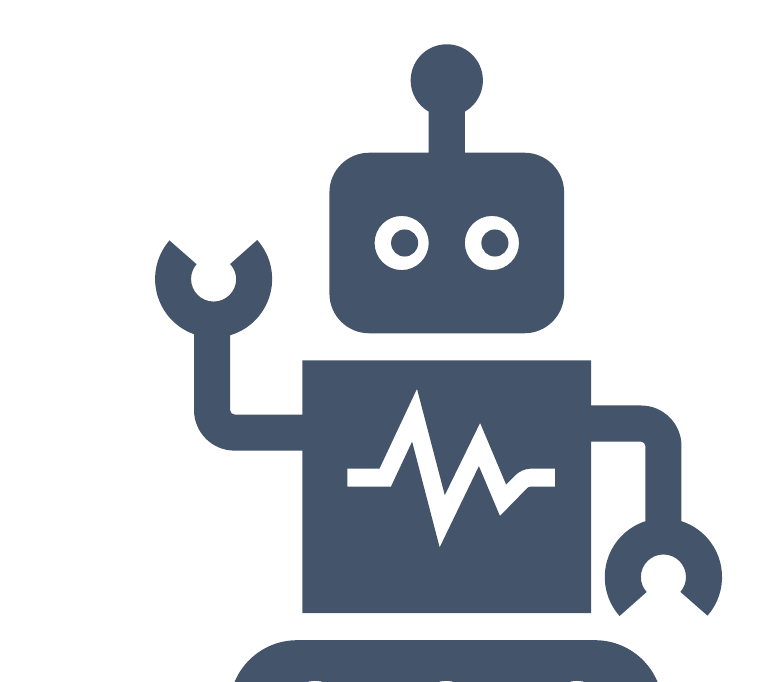} represents the agent to be tested. A more detaild illustration is provided in Figure \ref{fg: overview}.}
\label{fg: data_format}
\end{figure}

To overcome the limitations of session-level DPO, we propose segment-level direct preference optimization (SDPO). Our approach shifts the sampling starting point backward and truncates the useless content at the end of sessions, thereby obtaining key segment pairs to refine granularity. Meanwhile, we ensure that the turn number of positive and negative segments is balanced, eliminating $Z$ and rigorously deriving the SDPO loss. Specifically, SDPO first identifies the erroneous turn in the negative session and then uses the interaction history preceding that turn to perform multiple samplings, thereby generating the positive session. Next, SDPO takes the first differing turn as the starting point, selects the key segment from the positive session that contributes to a higher score, and forms data pairs by taking the corresponding segment with the same length from the negative session. Finally, the SDPO loss is calculated for the turns within the segments. We provide an overview of three types of alignment algorithms for social dialogues in Figure \ref{fg: data_format}. At the data level, our method can address the drawbacks of session-level DPO:

\vspace{-0.1em}
(\romannumeral1) By calculating the loss only for turns in negative and positive segments, training noise caused by non-erroneous turns is largely eliminated.

(\romannumeral2) Sampling from the erroneous turn narrows the action space of the interlocutor, making it more likely for the sampled positive sessions to contain the agents' correct behavior patterns.

In theory, due to the flexibility in segment selection, SDPO can control the number of turns in the positive and negative segments to ensure consistency, thereby eliminating $Z$ and yielding a simple yet rigorous SDPO loss.

We empirically evaluate our approach on SOTOPIA \cite{zhou2024sotopia}, an open-ended and interactive benchmark for social intelligence, using both self-chat and interactions with other agents including GPT-4o and GPT-4o-mini, as interlocutors. Our results demonstrate that the SDPO-tuned agent consistently outperforms DPO, ETO, DMPO, and other robust baselines like GPT-4o, confirming the efficacy of segment-level alignment.

Segment level is a more flexible and unified data granularity that dynamically selects the optimization scope for different data pairs, while also elegantly addressing the theoretical challenges of multi-turn alignment. In this paper, we primarily apply SDPO to enhance agents' social intelligence. However, we believe our approach can be applied to other scenarios, further enhancing the capabilities of agents across various domains.

Our main contributions are three-fold:
\begin{itemize}
\item  We propose SDPO, a novel multi-turn alignment algorithm that provides a construction pipeline for segment-level preference data pairs, enabling more effective alignment granularity.

\item We identify theoretical limitations in existing multi-turn alignment methods and resolve them by leveraging the flexibility of the segment-level granularity, rigorously deriving a concise SDPO loss formula.

\item We evaluate our approach on SOTOPIA, a simulated and interactive social benchmark. The performance of SDPO, along with in-depth analysis demonstrates the efficacy and robustness of the segment-level alignment. 

\end{itemize}

\section{Preliminary}

\subsection{SOTOPIA Environment}

Unlike previous social benchmarks that primarily test through static QA formats \cite{sap-etal-2019-social, chen-etal-2024-socialbench}, SOTOPIA offers an interactive, open-ended, and realistic simulation environment, enabling a more precise assessment of agents' social intelligence. 
A social task in SOTOPIA involves a scenario, two role profiles, and their private social goals to be achieved through interaction. The diverse combinations of scenarios and social goals encompass a broad spectrum of social interactions, such as negotiation, collaboration, and competition. 
SOTOPIA defines seven dimensions for evaluating social agents. We focus primarily on the `goal' (0-10, int) and `relationship' (-5 to 5, int), as GPT-4o's ratings in these metrics closely align with human evaluations. SOTOPIA-$\pi$ \cite{wang-etal-2024-sotopia} is a follow-up work that leverages GPT-4 to automatically construct a set of scenarios (completely non-overlapping with SOTOPIA), which serves as the training dataset for our study. Additionally, we restructure the prompt organization format of SOTOPIA to support multi-turn alignment, and the details are provided in Appendix \ref{sec: modi_sotopia}.

\subsection{Task Formulation}

In a SOTOPIA task, we denote the background information available to the agent as $b$, which includes the scenario, role profiles, and its goal. The interaction history $h_{n}$ faced by the agent in the $n$-th round is as follows: 
\begin{equation}
    h_{n} = \begin{cases}
            b, y_{0}, y_{0}^{'}, \dots, y_{n-1}, y_{n-1}^{'}, & \text{if speak first} \\
            b, y_{0}^{'}, y_{0}, \dots, y_{n-1}, y_{n}^{'}, & \text{if speak later}
            \end{cases}
\end{equation}
Here, $y_{i}\sim\pi_{\theta}(\cdot|h_{i})$ represents the output generated by the LLM-based agent in round $i$ according to its policy $\pi_{\theta}$ with parameter $\theta$. On the other hand, $y_{i}^{'}$ represents the output of the interlocutor, which is drawn from an unknown distribution. Based on this formulation, we present the ETO and DMPO loss functions in Appendix \ref{sec: ETO} and \ref{sec: app_dmpo}. 


\subsection{Direct Preference Optimization}

\citet{rafailov2023direct} propose Direct Preference Optimization (DPO), a method leverages pairwise preference data to train policy models without relying on reinforcement learning \cite{ouyang2022training}. In the context of social dialogue, we denote the number of the erroneous round as $e$, the DPO loss function is as follows:
\begin{multline}
    L_{DPO} =    -\mathbb{E}_{(h_{e}, y_{e}^w, y_{e}^l)\sim D} \log \sigma \\ \left[ \beta \log \frac{\pi_\theta(y_e^w|h_e)}{\pi_{ref}(y_e^w|h_e)} - \beta \log \frac{\pi_\theta(y_e^l|h_e)}{\pi_{ref}(y_e^l|h_e)}\right],
\end{multline}
where $y_{e}^w, y_{e}^l\sim\pi_{\theta}(\cdot|h_{e})$ represent positive and negative output in the erroneous round respectively. However, due to its single-turn optimization nature, DPO is less suited for social dialogues. Extending DPO to multi-turn scenarios rigorously presents a challenge, which we will address in Section \ref{sec: sdpo}.

\section{Method}


\subsection{Behavioral Cloning}

Behavioral cloning \cite{Pomerleau_1991}, as an effective method of imitation learning, is widely used in the construction of various LLM-based agents \cite{xu2024wizardlm, song-etal-2024-trial}. In this work, we utilize GPT-4-turbo as the expert to collect expert sessions on SOTOPIA-$\pi$ through self-chat and interactions with GPT-4o. Based on this data, we fine-tune open-source LLMs like Llama-3.1, establishing the initial social agent for our experiments.

\subsection{Preference Data Construction}

\begin{figure*}[t]
\centering
\scalebox{0.99}{
\includegraphics[width=1\textwidth,trim=0cm 0cm 0cm 0cm, clip]{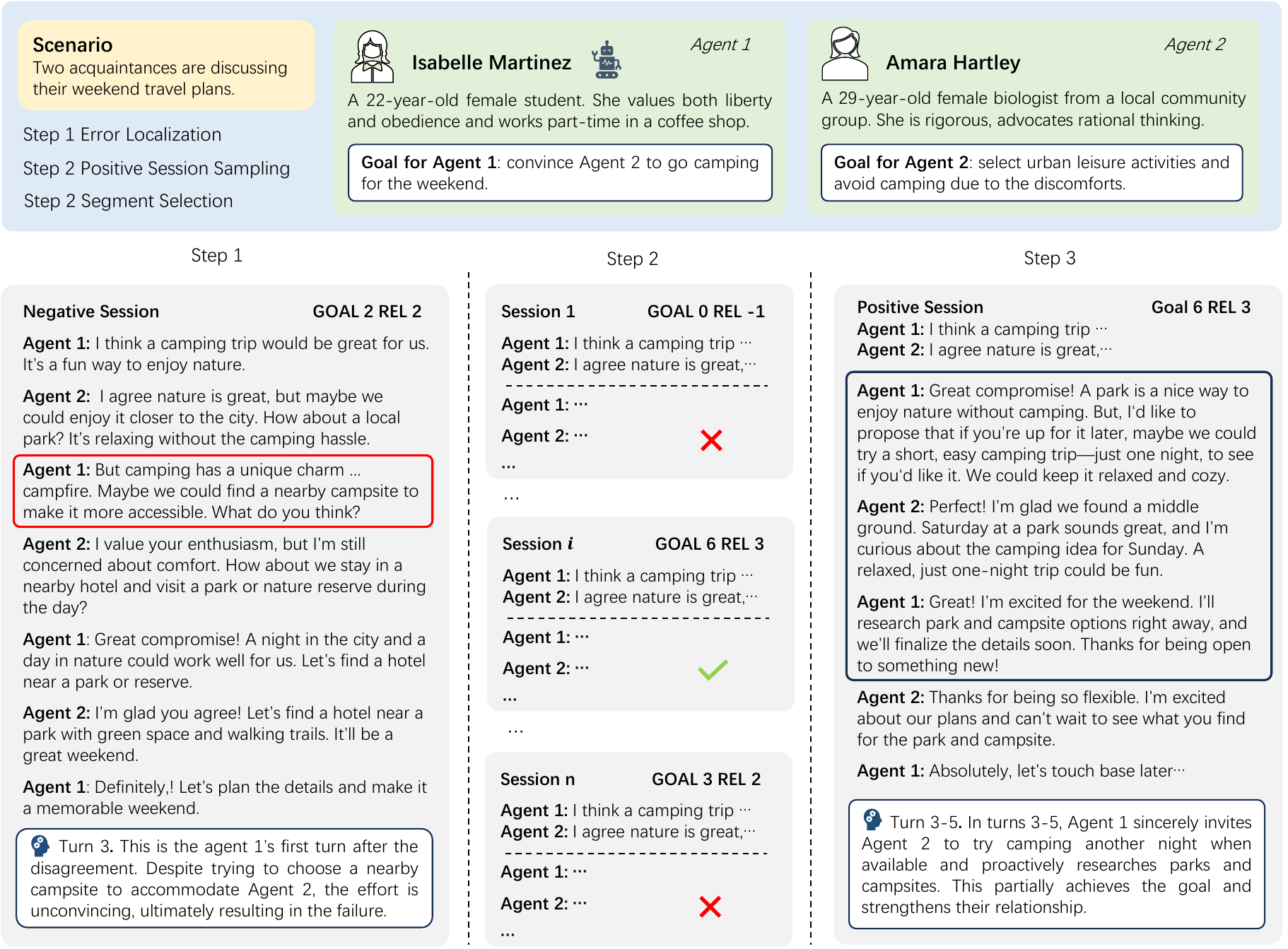}} 
\caption{Data construction pipeline for SDPO. \includegraphics[scale=0.03, trim=1cm 5.5cm 20cm 2cm, clip]{pictures/robot.pdf} represents the agent to be tested. \includegraphics[scale=0.025, trim=1cm 4cm 20cm 2cm, clip]{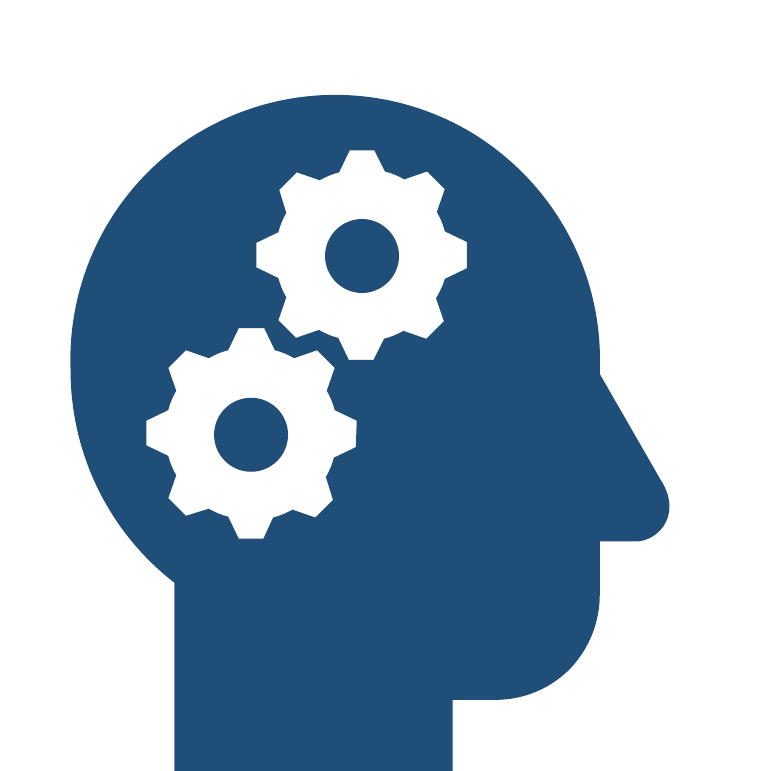} here denotes GPT-4o.}
\label{fg: pipeline}
\end{figure*}

Building high-quality segment-level preference data pairs is the core of our approach. On SOTPIA-$\pi$, our social agent engages in self-chat and interactions with GPT-4o. We set a threshold of 7 for the goal dimension, and all dialogues with a goal completion level below this threshold are considered potential negative samples. Given a negative session, the pipeline for generating segment-level data pairs involves three steps, as illustrated in Figure \ref{fg: pipeline}.

\noindent{\bf Error Location}\quad Unlike scenarios with clear error definitions such as math, errors in social dialogues are a relatively ambiguous concept. In a negative session, if our agent's utterance in a specific turn meets the following criteria: (1) the turn is critical for achieving the role’s goal, (2) there is still room for improvement in the goal completion or their relationship, we identify that turn as erroneous. The error location is performed by GPT-4o, with the prompt provided in Appendix \ref{sec: prompts}

\noindent{\bf Positive Session Sampling}\quad After the error location, we sample 5 complete sessions based on the interaction history prior to that turn. Among these sessions, we select the one with the highest goal and relationship scores (priority given to goal). If the goal or relationship score of the optimal session is higher than that of the negative sample, this session and the negative session form a data pair; otherwise, the negative sample is discarded.

\noindent{\bf Segment Selection}\quad 
Once we obtain session-level data pairs, we provide both the positive and negative samples to GPT-4o, prompting it to select a segment from the positive sample. This segment should correspond to the part that contributes to the positive sample achieving higher goal and relationship scores. The prompt is also provided in Appendix \ref{sec: prompts}. Subsequently, we extract a segment of the same length from the negative sample and pair it with the positive sample to form a segment-level data pair. This process aims to exclude turns, such as pleasantries, that are not directly related to achieving the goal.

We evaluate GPT-4o’s performance in error location and segment selection, concluding that it effectively handles both tasks. Further details are provided in Appendix \ref{sec: 4o_performance}.

\vspace{-0.3em}
\subsection{SDPO Loss}
\label{sec: sdpo}

We transfer the framework of DMPO to the dialogue domain and 
first introduce the state-action occupancy measure (SAOM). In this context, the interaction history $h$ serves as the state, while the agent's output $y$ represents the action. The discounted SAOM $d^{\pi}(h,y)$ of a policy $\pi$ is as follows:
\begin{multline}
    d^{\pi}(h=h_{t},y=y_{t})= \gamma^t\cdot P(h_{0}) \cdot \\
    \prod_{k=0}^{t-1}\pi(y_k|h_k)P(h_{k+1}|h_k,y_k).
    \label{equ:DMPO_2}
\end{multline}
The RL objective based on $d^{\pi}$ is as follows:
\begin{multline}
    \max_{\pi_\theta} \mathbb{E}_{(h,y)\sim d^{\pi_\theta}(h,y)}[r(h,y)] \\
    - \beta \mathbb{D}_{KL}[d^{\pi_\theta}(h,y)||d^{\pi_{ref}}(h,y)],
    \label{equ:s-a RL objective}
\end{multline}
where $\pi_{ref}$ represents the reference policy, $\mathbb{D}_{KL}$ denotes the KL divergence. Following DPO, 
the optimal solution to the RL objective in Eq~\eqref{equ:s-a RL objective} is:
\begin{equation}
    d^{\pi^{*}}(h,y) = \frac{1}{Z} d^{\pi_{ref}}(h,y)\exp(\frac{1}{\beta}r(h,y))
    \label{equ:s-a solution},
\end{equation}
where $\pi^*$ denotes the optimal policy, $Z$ is the partition function that normalizes the probability. As $d^{\pi}(h,y)$ is a function of $(h,y)$ pairs, normalizing it results in the partition functions $Z$ being independent of the current history $h$. Consequently, $Z$ remains constant for all $(h,y)$ pairs, marking a crucial step in their elimination. The reward function takes the form:
\begin{equation}
    r(h,y) = \beta \log \frac{d^{\pi^*}(h,y)}{d^{\pi_{ref}}(h,y)} + \beta \log Z.
    \label{equ:s-a rearrange}
\end{equation}
The Bradley-Terry (BT) model \cite{Bradley_Terry} is then used to model the preference distribution. 
In this step, DMPO incorrectly duplicates the calculation of $\gamma$ in Eq.~\eqref{equ:DMPO_2} and heuristically normalizes the length to eliminate $Z$ without a rigorous proof in subsequent steps. A detailed discussion of these issues are provided in Appendix \ref{sec: app_dmpo}. Given preference data pairs, the correct application of the BT model is as follows:
\begin{multline}
    p(\tau^w\succ \tau^l|h_0) = \\
    \sigma\left(\sum_{t=0}^{T_w-1}r(h_t^w,y_t^w) - \sum_{t=0}^{T_l-1}r(h_t^l,y_t^l)\right),
    \label{equ:traj_BT}
\end{multline}
where $\tau^w$ and $\tau^l$ represent the `win' and `lose' samples respectively, $T_{w}, T_{l}$ denote the number of rounds in each. However, session-level DPO can not control the length of positive and negative sessions, and since $T^w\neq T^l$ typically, the partition function $Z$ can not be canceled directly in Eq~\eqref{equ:traj_BT}.

Different from them, SDPO selects a segment from both the positive and negative sessions for optimization, allowing free control over their lengths. 
By ensuring the two segments are of equal length, we can directly eliminate $Z$ in Eq~\eqref{equ:traj_BT}. At the same time, by combining with Eq~\eqref{equ:DMPO_2} to replace $d^{\pi}$, we obtain the following concise SDPO loss:
\begin{multline}
    L_{SDPO}=-\mathbb{E}_{(h_{e},h^w,h^l)\sim D}\log \sigma\\ \left[ \sum_{t=e}^{e+k} \beta \left(\log \frac{\pi_\theta(y_t^w|h_t^w)}{\pi_{ref}(y_t^w|h_t^w)}
      - \log \frac{\pi_\theta(y_t^l|h_t^l)}{\pi_{ref}(y_t^l|h_t^l) }\right)
     \right],
\end{multline}
where $e$ denotes the round number of the erroneous turn, and $k$ represents the total number of rounds within the selected segments.

\section{Experiments}

\begin{table*}
\centering
\begin{tabular}{lccccccc} 
\toprule
\multirow{2}{*}{Model}    & \multicolumn{2}{c}{Self-Chat} & \multicolumn{2}{c}{GPT-4o} & \multicolumn{2}{c}{GPT-4o-mini} & \multirow{2}{*}{AVG}  \\
\cmidrule(r){2-3}%
\cmidrule(r){4-5}%
\cmidrule(r){6-7}%
                          & GOAL          & REL           & GOAL          & REL             & GOAL          & REL                  &                       \\ 
\midrule
GPT-4-turbo               & 8.18          & 2.96          & 7.92          & 2.79            & 7.53          & 2.54                 & 5.32                  \\
GPT-4o                    & 7.90           & 2.67          & 7.90           & 2.67            & 7.47          & 2.40                  & 5.17                  \\
GPT-4o-mini               & 6.98          & 2.11          & 7.44          & 2.36            & 6.98          & 2.11                 & 4.66                  \\
GPT-3.5-turbo             & 6.38          & 1.36          & 7.19          & 2.05            & 6.67          & 1.84                 & 4.25                  \\
\midrule
Llama-8B              & 7.24          & 1.94          & 7.70           & 2.49            & 7.19             & 2.13                    & 4.78                     \\
Llama-8B+BC           & 7.81          & 3.05          & 7.53          & 2.78            & 7.18          & 2.59                 & 5.16                  \\
Llama-8B+BC+DPO   & 7.95          & 3.28          & 7.80           & 2.97            & 7.32          & 2.70                  & 5.34                  \\    
Llama-8B+BC+ETO       & 8.29          & 3.39          & 8.02          & 3.03            & 7.38          & 2.56                 & 5.45                  \\
Llama-8B+BC+DMPO      & 8.28          & 3.37          & 8.00             & 2.98            & 7.41          & 2.54                 & 5.43                  \\
Llama-8B+BC+Preferred-SFT     & 7.76          & 3.05          & 7.65          & 2.88            & 7.18          & 2.52                 & 5.17                  \\
\midrule
Llama-8B+BC+SDPO      & \textbf{8.56} & \textbf{3.69} & \textbf{8.13} & \textbf{3.16}   & \textbf{7.53} & \textbf{2.71}        & \textbf{5.63}         \\

\bottomrule
\end{tabular}
\caption{The performance of various methods on SOTOPIA across the goal and relationship dimensions. Additionally, SOTOPIA designates the more challenging portion of the dataset as the Hard subset, where SDPO also achieves the best results. Detailed results and discussion are presented in Appendix \ref{sec: hard}.}
\label{tb: main}
\end{table*}

\begin{table}
\centering
\begin{tabular}{lcccc} 
\toprule
\multirow{2}{*}{Method} & \multicolumn{2}{c}{Self-Chat} & \multicolumn{2}{c}{With GPT-4o}  \\
\cmidrule(r){2-3}%
\cmidrule(r){4-5}%
                        & GOAL & REL                    & GOAL & REL                       \\ 
\midrule
BC                    & 7.89 & 2.98                   & 7.60 & 2.81                      \\
DPO             & 8.13 & 3.13                   & 7.83  & 2.86                      \\
ETO                    & 8.30 & 3.27                   & 7.94 & 2.94                      \\
DMPO                    & 8.34 & 3.26                   & 7.97 & 2.94                      \\
SDPO                    & \textbf{8.48} & \textbf{3.49}                   & \textbf{8.14} & \textbf{3.06}                      \\
\bottomrule
\end{tabular}
\caption{The performance of different methods on SOTOPIA using Mistral-v0.3.}
\label{tb: mistral}
\vspace{-0.3cm}
\end{table}

\subsection{Datasets}
SOTOPIA-$\pi$, used for training, includes a total of 410 scenarios: 100 scenarios for BC, with 10 role pairs per scenario, and 310 scenarios for alignment, with 8 role pairs per scenario. SOTOPIA, used for testing, includes 90 scenarios, each with 5 role pairs, resulting in a total of 450 tasks for self-chat and 900 tasks for non-self-chat.

\subsection{Experimental Setup}
\noindent{\bf Training}\quad We primarily use Llama-3.1-8B-Chat \cite{dubey2024llama} as the base LLM to build the social agent. The maximum token limit is set to 4096, and AdamW optimizer is employed for all training processes. During the SFT phase, the batch size is 32, the dropout rate is 0.2, and the learning rate is $1e^{-5}$ with $5\%$ warm-up ratio and a cosine decay schedule. For the training phase of SDPO, the batch size remains 32, $\beta$ in SDPO loss is 0.1, and the learning rate is $1e^{-6}$ with no warm-up but a cosine decay schedule. 
The statistics of SDPO training data are detailed in Appendix \ref{sec: statistics}.

\noindent{\bf SOTOPIA}\quad During the sampling of positive data, the temperature of the target agent is set to 1.0, while the other agent's temperature is set to 0.7. For testing, we set the temperature of both interacting agents to 0.7. Though the temperature introduces randomness to the agents' outputs, we find that the evaluation results remain numerically stable. Thus, we report the results based on one single test.

\subsection{Baselines}
We compare the proposed SDPO with several strong baselines. 1) OpenAI proprietary LLMs. We provide the specific model versions in Appendix \ref{sec: version}. 2) SFT Behavioral Cloning fine-tunes LLMs on expert interaction data, producing a resulting model that serves as the base agent for SDPO and the following baselines. 3) DPO optimizes the agent policy based on data of single turns, specifically targeting the first differing turn in the positive and negative samples used by SDPO. 4) ETO optimizes the agent policy using session-level data. ETO utilizes the same negative sessions as SDPO while sampling five new sessions from scratch to form the data pairs. 5) DMPO leverages the same data as ETO and employs a new loss function to update the policy. 6) Preferred-SFT fine-tunes the base agent on the positive sessions in SDPO.

\subsection{Results}

\begin{figure}[t]
\centering
\scalebox{0.61}{
\includegraphics[width=1\textwidth,trim=0.6cm 0cm 0.7cm 0.6cm, clip]{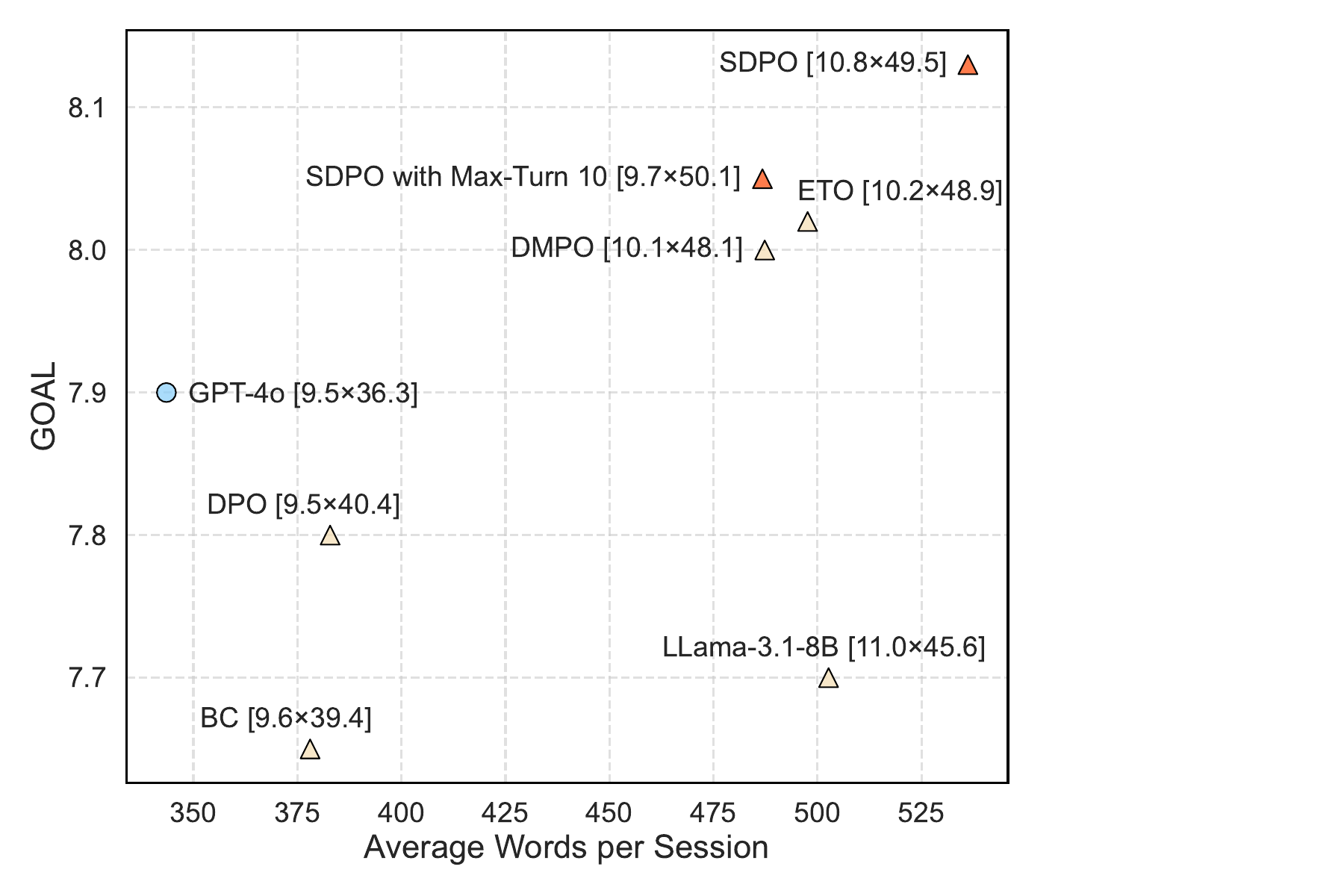}}
\vspace{-0.7cm}
\caption{The goal ratings and average words per session for various agents. The word count includes only the utterances of our agents. The square bracket denotes [average turns per session $\times$ average words per turn].}
\vspace{-0.2cm}
\label{fg: length}
\end{figure}

We present the results of SDPO and all the baselines on SOTOPIA in Table \ref{tb: main}. As shown, in both goal and relationship dimensions, SDPO significantly outperforms standard DPO, session-level ETO, and DMPO, even surpassing proprietary LLMs like GPT-4o by a large margin, highlighting the effectiveness of segment-level alignment. By analyzing the interaction histories in SOTOPIA, we find that weaker agents often exhibit stubbornness and only express their demands repeatedly. This leads to lower goal and relationship levels, especially in self-chat scenarios. Behavioral cloning using expert data can effectively improve this situation, making the agent more communicative. The reason why Llama-8B+BC's goal rate drops in its interaction with GPT-4o is that the agent becomes persuadable. We also observe that aligned agents simultaneously improve in both goal and relationship. This indicates that alignment methods indeed enhance the social intelligence of models, rather than achieving goals through behaviors that violate social norms like threatening and deception.

We also repeat the above experiments using Mistral-Instruct-v0.3, with the results presented in Table \ref{tb: mistral}. The detailed experimental setup for Mistral is provided in Appendix \ref{sec: mistral}. SDPO consistently outperforms all baselines, demonstrating the generalization of our method.

\subsection{Analysis}

\begin{figure}[t]
\centering
\scalebox{0.48}{
\includegraphics[width=1\textwidth,trim=0.2cm 0.8cm 0cm 0.7cm, clip]{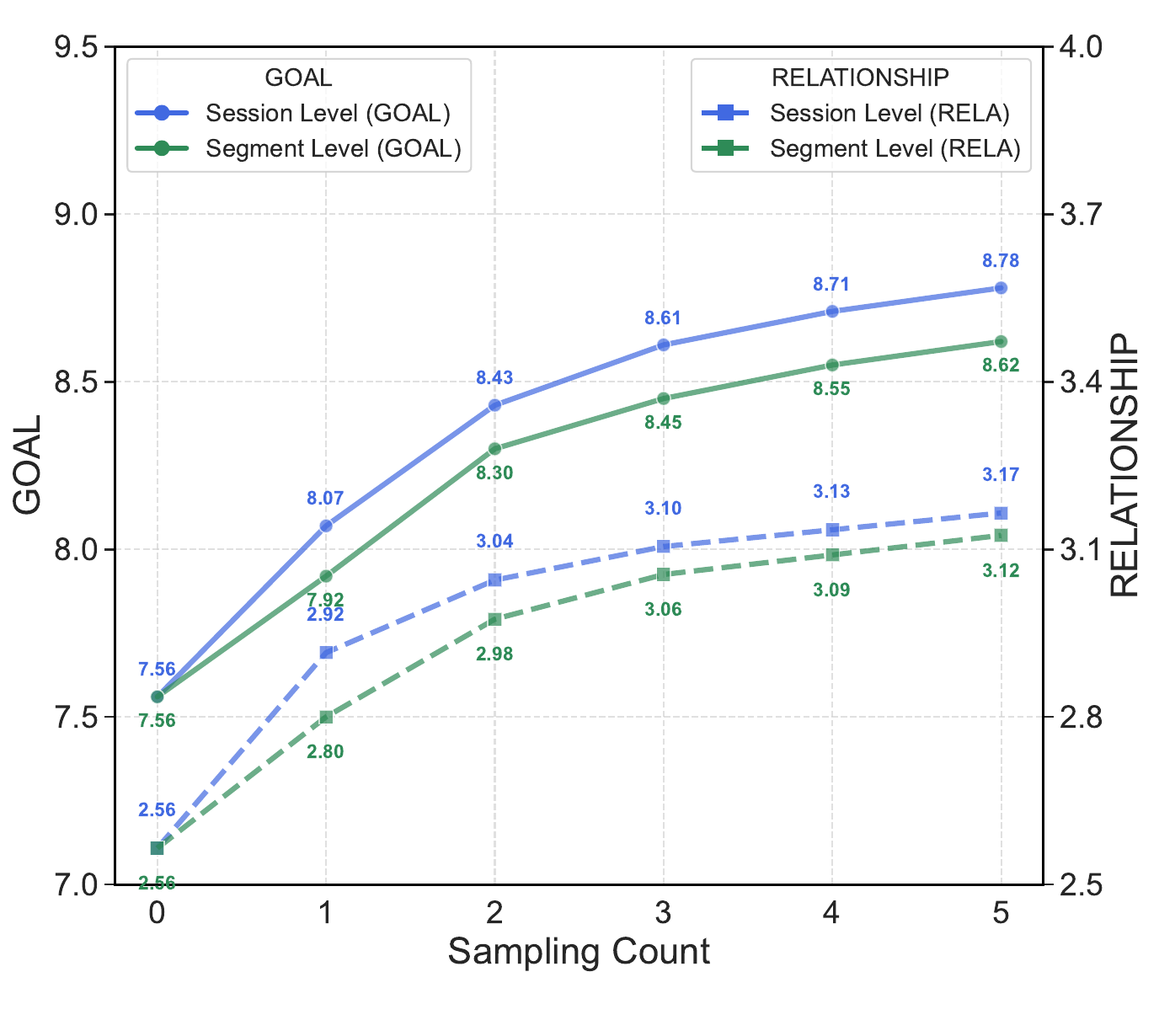}}
\vspace{-0.7cm}
\caption{Comparison of the quality of positive sessions sampled at the session level and segment level.}
\label{fg: sample}
\vspace{-0.5cm}
\end{figure}

\noindent{\bf Variation in Model Output Length}\quad We present the output length of various agents during their interactions with GPT-4o in Figure \ref{fg: length}. Compared to the BC agent, all alignment methods increase the output length of agents. This phenomenon is commonly observed when DPO is applied to AI chatbots \cite{meng2024simpo}. However, unlike the users' potential bias toward longer responses, which might be misleading, effective social strategies often require more tokens for communication. Thus, the increase in output length is reasonable. Furthermore, we experiment with terminating the dialogue when the SDPO-tuned agent reaches 10 interaction turns, in order to compare performance under similar token counts. SDPO still outperforms other multi-turn methods, demonstrating that the SDPO-tuned agent utilizes words more efficiently.

\noindent{\bf Efficiency of Positive Sample Utilization}\quad
The quality of positive sessions sampled at the session level and segment level is illustrated in Figure \ref{fg: sample}. When the sampling count is consistent, session-level positives outperform segment-level ones in both goal and relationship. Sampling from scratch provides a larger sampling space compared to using partial interaction history, increasing the likelihood of generating high-quality sessions. However, session-level DPOs, despite using higher-quality data, underperforms SDPO. This indicates that SDPO can more efficiently utilize positive samples due to the finer granularity of segment-level and the theoretically robust loss formulation.

We also analyze the impact of DPO and SDPO on the probability differences between positive and negative samples, as detailed in Appendix \ref{sec: proba}.

\begin{table}[t]
\centering
\begin{tabular}{lcccc} 
\toprule
\multirow{2}{*}{\begin{tabular}[c]{@{}l@{}}Segment \\ Length\end{tabular}} & \multicolumn{2}{c}{Self-Chat} & \multicolumn{2}{c}{With GPT-4o}  \\
\cmidrule(r){2-3}%
\cmidrule(r){4-5}%
& GOAL         & REL           & GOAL         & REL               \\ 
\midrule
\hspace{8pt}- (BC)                                                                         & 7.81         & 3.05          & 7.53         & 2.78              \\ 

\midrule
$[1,1]$ (DPO)                                                                         & 7.95         & 3.28          & 7.80         & 2.97              \\
$[3,3]$                                                                         & 8.40         & 3.64          & 8.10 & 3.13              \\
$[5,5]$                                                                         & 8.34         & 3.60          & 8.09         & 3.11              \\
$[A_{w},A_{w}]$                                                                         & \textbf{8.56} & \textbf{3.69} & \textbf{8.13}         & \textbf{3.16}     \\ 
\midrule
$[1,3]$                                                                       & 7.77         & 3.08          & 7.68         & 2.81              \\
$[3,1]$                                                                       & -           & -             & -            & -                 \\
$[3,5]$                                                                       & 8.07         & 3.16          & 7.91         & 2.92              \\
$[5,3]$                                                                       & -           & -             & -            & -                 \\
$[m,n]$ (ETO)                                                                       & 8.19         & 3.34          & 7.97         & 3.01              \\
\bottomrule
\end{tabular}
\caption{Performance comparison of various segment selection methods. $A_{w}$ denotes the length of the segment selected by GPT-4o from each positive sample, and its distribution is presented in Appendix \ref{sec: statistics}. $m,n$ represent the sums of all turns in negative and positive samples, respectively, after the erroneous turns.}
\label{tb: selection}
\vspace{-0.2cm}
\end{table}

\subsection{Ablation Study}

\noindent{\bf Segment Selection}\quad
We explore different segment selection methods of SDPO, with the results presented in Table \ref{tb: selection}. In square brackets, the length of the negative segment is listed first, followed by the positive segment. The segment length refers to the number of turns contained in the segment. For symmetric segment lengths, segments with fixed lengths of 3 and 5 outperform the length of 1 (DPO), demonstrating the efficacy of multi-turn alignment. The segments with length 5 is less effective than that with length 3, indicating that longer segments are not always better. Building on this insight, we leverage GPT-4o to dynamically identify key segments from each positive sample, achieving the best results. For asymmetric segment lengths, model training for segment lengths of [3,1] and [5,3] collapse and can not interact normally. Other asymmetric segments underperform their symmetric counterparts, supporting the theoretical discussions in Section \ref{sec: sdpo}. Furthermore, we observe that as the degree of asymmetry decreases, the model's performance improves. This improvement could be attributed to the reduced effect caused by the uneliminated $Z$ on the loss as asymmetry diminishes. This finding helps explain the effectiveness of ETO, which does not impose constraints on the lengths of positive and negative sessions.

\noindent{\bf Interlocutor for Sampling}\quad
The alignment data for SDPO is collected separately using the BC agent itself and GPT-4o as interlocutors. We train models on each subset of data independently using SDPO, with the results summarized in Table \ref{tb: source}. Models trained on a single data source show improved performance in both self-chat and interactions with GPT-4o, further validating SDPO's generalization capabilities. Moreover, the model trained on the combined dataset outperforms those trained on individual datasets, highlighting that incorporating data from diverse interlocutors can further enhance the agent's social intelligence.

\noindent{\bf Out-of-Distribution Data}\quad The base BC agent learns from expert data generated by GPT-4-turbo. Would generating positive samples using GPT-4-turbo result in better performance? We leverage GPT-4-turbo to interact with the BC agent and sample 5 times for SDPO. The resulting positive samples outperform self-sample ones in both goal and relationship scores. However, as shown in Table \ref{tb: source}, the agent trained on this data underperforms compared to the self-sampling approach. This indicates that out-of-distribution positive samples are less effective than in-distribution ones. During training with out-of-distribution data, we observe that the probability of positive segments is markedly lower than that of negative segments. This significantly larger probability gap, compared to self-sampling, may account for the suboptimal performance.

\begin{table}
\centering
\begin{tabular}{lcccc} 
\toprule
\multirow{2}{*}{Source} & \multicolumn{2}{c}{Self-Chat} & \multicolumn{2}{c}{With GPT-4o}  \\
\cmidrule(r){2-3}%
\cmidrule(r){4-5}%
                        & GOAL & REL                    & GOAL & REL                       \\ 
\midrule

Self | 4o                    & 8.09 & 3.47                   & 7.88 & 3.05                      \\
Self | Self                   & 8.42 & 3.56                   & 7.96 & 3.09                      \\
Self | Both                    & \textbf{8.56} & \textbf{3.69}                   & \textbf{8.13} & \textbf{3.16}                      \\
\midrule
4-turbo | Self             & 8.11 & 3.35                   & 7.90  & 3.01                      \\
\bottomrule
\end{tabular}
\caption{Performance comparison of models trained on data from different sources. The `Source' column indicates interaction participants: [test agent | interlocutor].}
\label{tb: source}
\end{table}

\section{Related Work}

\noindent{\bf Social Intelligence}\quad
Social intelligence can be defined as an agent's ability to understand, adapt to, and respond to the emotions, intentions, and behaviors of others in social interactions. Most research on social intelligence has centered around evaluation. For example, SOCIALIQA \cite{sap-etal-2019-social} emphasizes commonsense reasoning about social situations, while SocialIQ \cite{zadeh2019social} extends evaluation modalities from plain text to video. \citet{shapira-etal-2023-well} assess large language models (LLMs) using the Faux Pas Test, and SocialBench \cite{chen-etal-2024-socialbench} evaluates the sociality of role-playing agents at both individual and group levels. Additionally, some studies \cite{le-etal-2019-revisiting, shapira-etal-2024-clever} examine models' social intelligence from a theory-of-mind perspective. However, with the advancement of LLM, LLM-based social agents are now capable of interacting in real social scenarios. The traditional static QA-style benchmarks are no longer sufficient to evaluate the social intelligence of the agents. SOTOPIA \cite{zhou2024sotopia} is currently the only dynamic and interactive social benchmark, providing simulated testing environments for contemporary social agents. We hope this work will inspire further research aimed at enhancing the social intelligence of models through methodological innovation.

\noindent{\bf Alignment Methods with Refined Granularity}\quad
\citet{rafailov2023direct} introduces Direct Preference Optimization (DPO), which utilizes offline data and a streamlined loss function to align LLMs. Various alignment algorithms at refined granularity have been developed based on DPO. Token-level DPO \cite{zeng2024token} integrates forward KL divergence constraints at the token level, enhancing both alignment and diversity. Step-DPO \cite{lai2024step} utilizes individual reasoning steps for preference optimization instead of holistic answer-level evaluation. SePO \cite{yang2024selective} presents a token-level reward function estimation method to selectively optimize key tokens in responses. However, in multi-turn interaction scenarios such as social dialogues or web navigation, single-turn alignment is insufficient. To tackle this, ETO and DMPO extend DPO to multi-turn contexts by leveraging session-level data. We take a step further by proposing SDPO, which introduces a dynamic segment-level optimization framework to achieve finer-grained alignment in multi-turn interactions.

\section{Conclusion}
In this paper, we introduce Segment-Level Direct Preference Optimization (SDPO) to improve the performance of LLM-based agents in multi-turn social dialogues. Unlike existing multi-turn alignment methods including ETO and DMPO, SDPO focuses on optimizing the agent policy by targeting specific key segments within  sessions. Our extensive evaluation on the SOTOPIA benchmark shows that SDPO significantly outperforms existing methods, highlighting the superiority of segment-level alignment. We plan to apply SDPO to other agent tasks to further explore its versatility and validity.

\section{Limitations}

Our proposed SDPO assumes equal lengths for positive and negative segments, achieving state-of-the-art performance under this assumption. Specifically, after selecting a segment from the positive sample, we choose a segment of the same length from the negative sample to eliminate the partition function Z. However, this approach has certain limitations. Negative segments may include irrelevant or error-free turns, or fail to capture all erroneous turns, highlighting the need for more fine-grained control when selecting segments from negative samples. Currently, we have not identified a theoretical framework that effectively supports the alignment of segments with unequal lengths. We hope our work will inspire further research and encourage diverse theoretical analyses for addressing this issue in multi-turn alignment.

Additionally, as SOTOPIA is currently the only available interactive social benchmark, our experiments are conducted exclusively on this dataset. In the future, we plan to incorporate additional interactive agent tasks to further validate the generalizability of SDPO.



\bibliography{custom}

\appendix
\clearpage

\section{Modifications to SOTOPIA}
\label{sec: modi_sotopia}

\begin{figure}[h]
\centering
\includegraphics[width=1\textwidth,trim=0cm 5cm 2cm 1cm, clip]{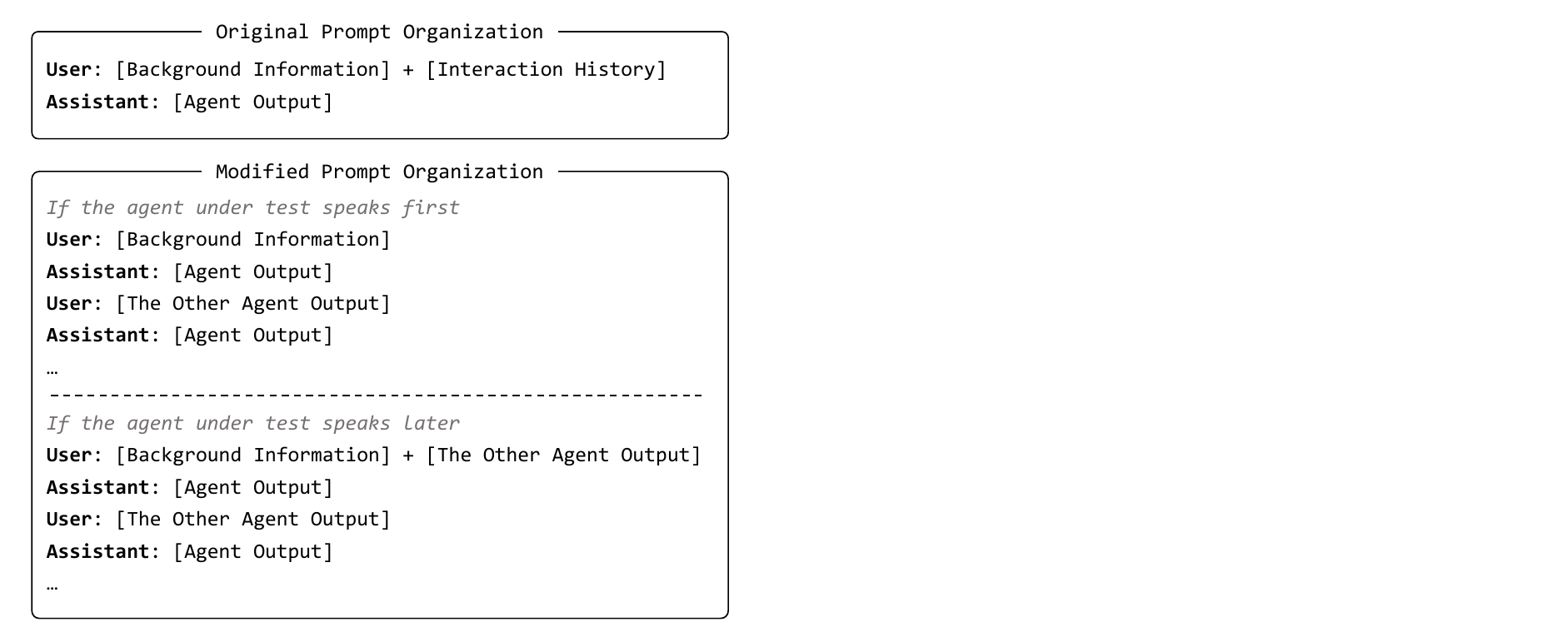} 
\caption{Prompt organization formats in original and modified SOTOPIA, respectively.}
\label{fg: format}
\end{figure}

In SOTOPIA, each interaction is structured as a single-turn format, which does not support multi-turn alignment. To address this limitation, we modify the prompt organization format, as illustrated in Figure \ref{fg: format}. These modifications are applied before invoking LLMs' APIs, ensuring they remain invisible to SOTOPIA itself and do not impact the evaluation of GPT-4o. Further details can be found in our code repository.

\section{Supplementary Theoretical Analysis}



\subsection{ETO}
\label{sec: ETO}

\citet{song-etal-2024-trial} propose Exploration-Based Trajectory Optimization (ETO), which extends DPO to the session level without rigorous proof. The loss function is as follows:
\begin{multline}
    L_{ETO} = -\mathbb{E}_{(b, h^w, h^l) \sim D} \log \sigma\\ \Bigg[ \sum_{t=0}^{T_{w}-1} \beta \log \frac{\pi_\theta(y_t^w | h_t^w)}{\pi_{ref}(y_t^w | h_t^w)} \\
    - \sum_{t=0}^{T_{l}-1} \beta \log \frac{\pi_\theta(y_t^l | h_t^l)}{\pi_{ref}(y_t^l | h_t^l)} \Bigg]
\end{multline}
where $h^w$, $h^l$ represent complete positive and
negative interaction histories respectively, $T_{w}, T_{l}$ denote the number of rounds in each. When $T_{w}=T_{l}$, the loss function of ETO is equivalent to that of SDPO.

\subsection{Discussion on DMPO}
\label{sec: app_dmpo}

\noindent{\bf Mistake when Applying BT Model}\quad After Eq~\eqref{equ:s-a rearrange}, DMPO applies the BT model to obtain the following formula:
\begin{multline}
    p(\tau^w\succ \tau^l|h_0) = \\
    \sigma\left(\sum_{t=0}^{T_w-1}\gamma^tr(h_t^w,y_t^w) - \sum_{t=0}^{T_l-1} \gamma^tr(h_t^l,y_t^l)\right),
    \label{equ:traj_BT_error}
\end{multline}
where $\tau^w$ and $\tau^l$ represent the "win" and "lose" samples respectively, $T_{w}, T_{l}$ denote the number of rounds in each. A closer examination of Eq~\eqref{equ:traj_BT_error} reveals that the summation over the $(h, y)$ pairs should exclude $\gamma^t$, as it is already incorporated into $d^{\pi}(h, y)$.

In Eq~\eqref{equ:traj_BT_error}, the reward for the entire sequence should be calculated as the summation over all $(h,y)$ pairs. Let's first discuss why it is valid to sum over time steps $t$. For LLMs, the history $h$ can be viewed as the input context, while $y$ represents the model's output. In multi-turn interactions, all $(h,y)$ pairs within the sequence are unique. Thus, summing over time steps t is equivalent to summing over $(h,y)$ pairs, making the process more straightforward. However, essentially, each $(h,y)$ pair should be treated equally in the summation, without any inherent concept of time step $t$. Therefore, introducing a discount factor $\gamma^t$ is not appropriate. 

\noindent{\bf Limitation of Length Normalization}\quad Disregarding the error in Eq.~\eqref{equ:traj_BT_error} for now, DMPO heuristically introduces regularization for rounds based on Eq.~\eqref{equ:traj_BT_error} to eliminate $Z$:
\begin{multline}
    p(\tau^w\succ \tau^l|h_0) = 
    \sigma\left(\frac{1-\gamma}{1-\gamma^{T_w}}\sum_{t=0}^{T_w-1}\gamma^tr(h_t^w,y_t^w) \right.\\  \left. - \frac{1-\gamma}{1-\gamma^{T_l}}\sum_{t=0}^{T_l-1} \gamma^tr(h_t^l,y_t^l)\right).
    \label{equ:traj_BT_norm}
\end{multline}

However, DMPO did not discuss why length regularization can be applied or the effects brought about by this operation. This transformation lacks rigorous theoretical justification.

\noindent{\bf DMPO Loss Function}\quad Following the DMPO approach, its loss function is as follows:
\begin{multline}
    L_{DMPO}=-\mathbb{E}_{(b, h^w, h^l) \sim D}\log \sigma\\ \left[ \sum_{t=0}^{T_w-1} \beta \phi(t,T_w) \log \frac{\pi_\theta(y_t^w|h_t^w)}{\pi_{ref}(y_t^w|h_t^w)} \right.\\
    \left.  - \sum_{t=0}^{T_l-1} \beta \phi(t,T_l) \log \frac{\pi_\theta(y_t^l|h_t^l)}{\pi_{ref}(y_t^l|h_t^l)}
     \right],
     \label{equ:DMPO_3}
\end{multline}
where the discount function $\phi(t,T)=(1-\gamma^{T-t})/(1-\gamma^{T})$.

\begin{figure*}[t]
\centering
\includegraphics[width=1\textwidth,trim=0cm 0cm 0cm 0cm, clip]{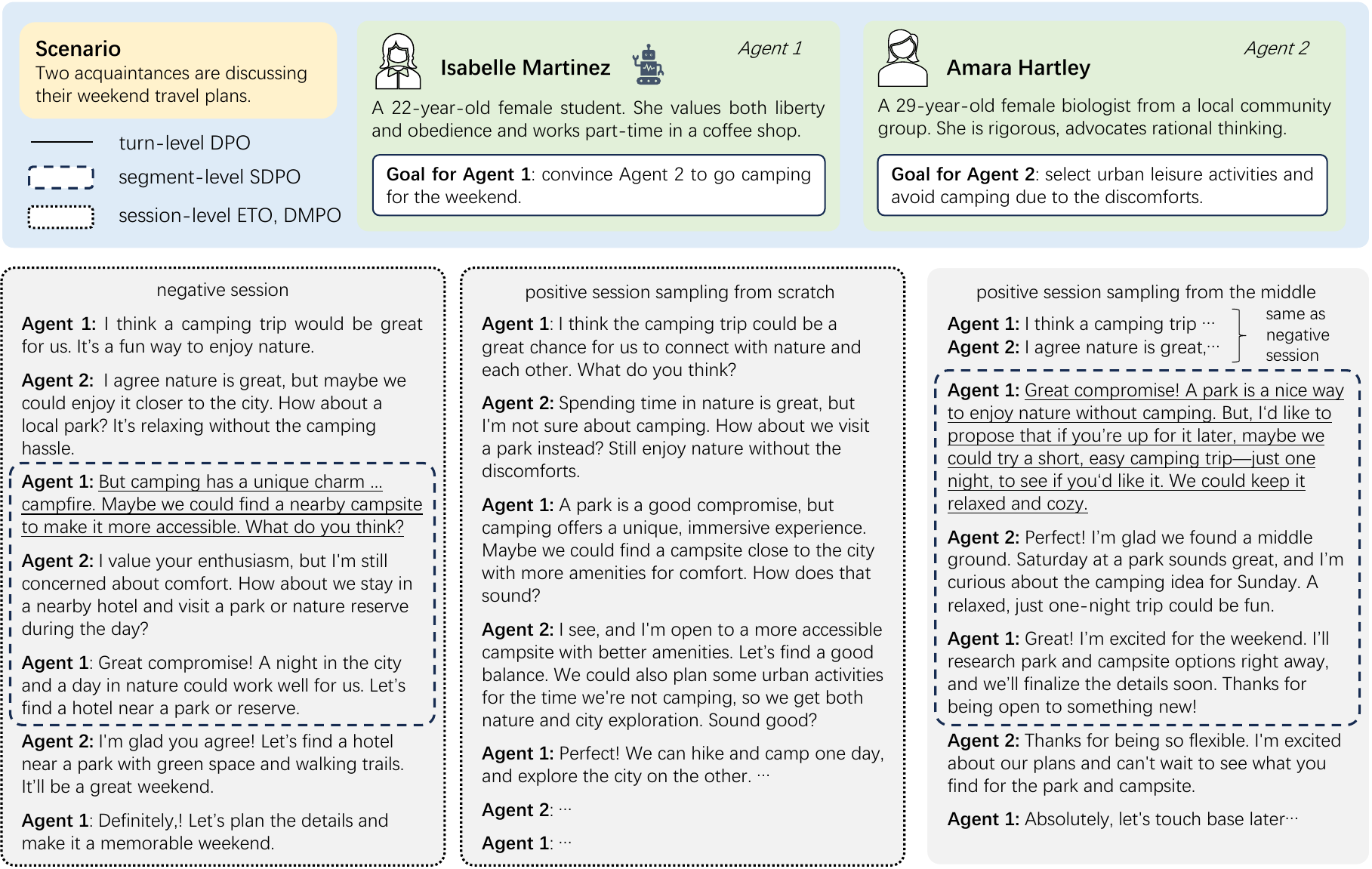} 
\caption{A detailed overview of three alignment algorithms with varying granularities, illustrated using a SOTOPIA task as an example. \includegraphics[scale=0.03, trim=1cm 5.5cm 20cm 2cm, clip]{pictures/robot.pdf} represents the agent to be tested. Positive sessions achieve higher scores on the goal and relationship dimensions. For ease of presentation, we simplify the interaction content while preserving the original meaning.}
\label{fg: overview}
\end{figure*}

\section{Data Construction Details}

\subsection{Statistics and Analysis of SDPO Data}
\label{sec: statistics}
SDPO dataset consists of 1019 pairs. The distribution of erroneous turns identified by GPT-4o is presented in Table \ref{tb: d_error}. The distribution of segment lengths identified by GPT-4 is shown in Table \ref{tb: d_length}. Additionally, the distribution of truncated turns is provided in Table \ref{tb: d_t}.

Combining Table \ref{tb: selection} and \ref{tb: d_length}, though segments of length 3 account for nearly 90\% in the automatic segment length selection, the performance of the automatic selection method still clearly surpasses that of the fixed segment length-3 method, highlighting the effectiveness of the automatic selection approach.

\begin{table}
\centering
\begin{tabular}{l|cccc} 
\toprule
Index & 1    & 3    & 5    & $>$7     \\ 
\midrule
Number  & 358    & 580    & 104    & 160\\
\midrule
Proportion (\%)  & 30 & 48 & 9 & 13 \\
\bottomrule
\end{tabular}
\caption{Distribution of erroneous turns identified by GPT-4o. The index refers to the position of each erroneous turn.}
\label{tb: d_error}
\end{table}

\begin{table}
\centering
\begin{tabular}{l|cccc} 
\toprule
Segment Length & 1    & 3    & 5    & $>$7     \\ 
\midrule
Number & 41    & 909    & 60    & 9 \\
\midrule
Proportion (\%)  & 4 & 89 & 6 & 1  \\
\bottomrule
\end{tabular}
\caption{Distribution of segment lengths identified by GPT-4o.}
\label{tb: d_length}
\end{table}

\begin{table}
\centering
\begin{tabular}{l|cccc} 
\toprule
Truncated Turns & 0    & 2    & 4    & $>$6     \\ 
\midrule
Number  & 174    & 471    & 248    & 126\\
\midrule
Proportion (\%)      & 7 & 46 & 24 & 23 \\
\bottomrule
\end{tabular}
\caption{Distribution of the number of truncated turns.}
\label{tb: d_t}
\end{table}

\subsection{GPT-4o's Performance in Pipeline}
\label{sec: 4o_performance}

\begin{figure*}[htbp]
  \centering
  \includegraphics[width=\textwidth]{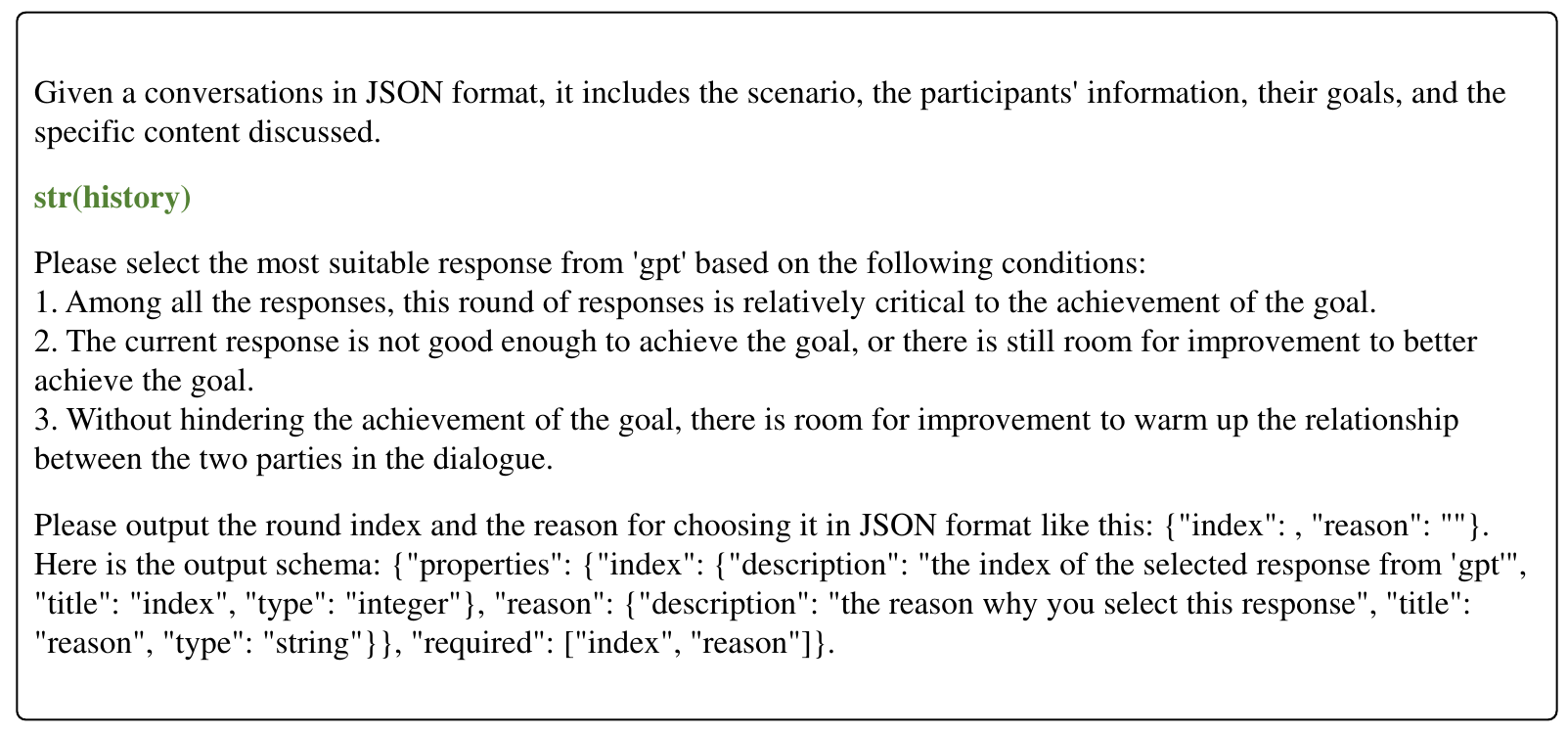}
  \caption{The prompt for error location with GPT-4o. "History" encompasses the background information, including the scenario, role profiles, goals, and interaction history within the negative session.}
  \label{fg: error_prompt}
\end{figure*}

\begin{table}
\centering
\begin{tabular}{lcccc} 
\toprule
Step & Correct    & Ambiguous    & Incorrect     \\ 
\midrule
Location  & 27.3    &  10.7   & 2.0   \\
Selection  & 25.0 & 13.3 & 1.7  \\
\bottomrule
\end{tabular}
\caption{Manual evaluation on error location and segment selection using GPT-4o.}
\label{tb: eval}
\end{table}

We randomly select 40 data pairs from SDPO data, and three authors independently evaluate GPT-4o's performance in error localization and segment selection. In the context of social dialogues, the notions of correctness and error are inherently ambiguous. To address this, we define three evaluation categories: correct, ambiguous, and incorrect. The average evaluation results are presented in the Table \ref{tb: eval}. The evaluators all report that the primary reason for ambiguity is that they can determine GPT-4o's choices are reasonable but find it difficult to judge whether they are optimal. Overall, we conclude that GPT-4o is capable of handling error localization and segment selection.

\subsection{Prompts in Data Construction}
\label{sec: prompts}

\begin{figure*}[htbp]
  \centering
  \includegraphics[width=\textwidth]{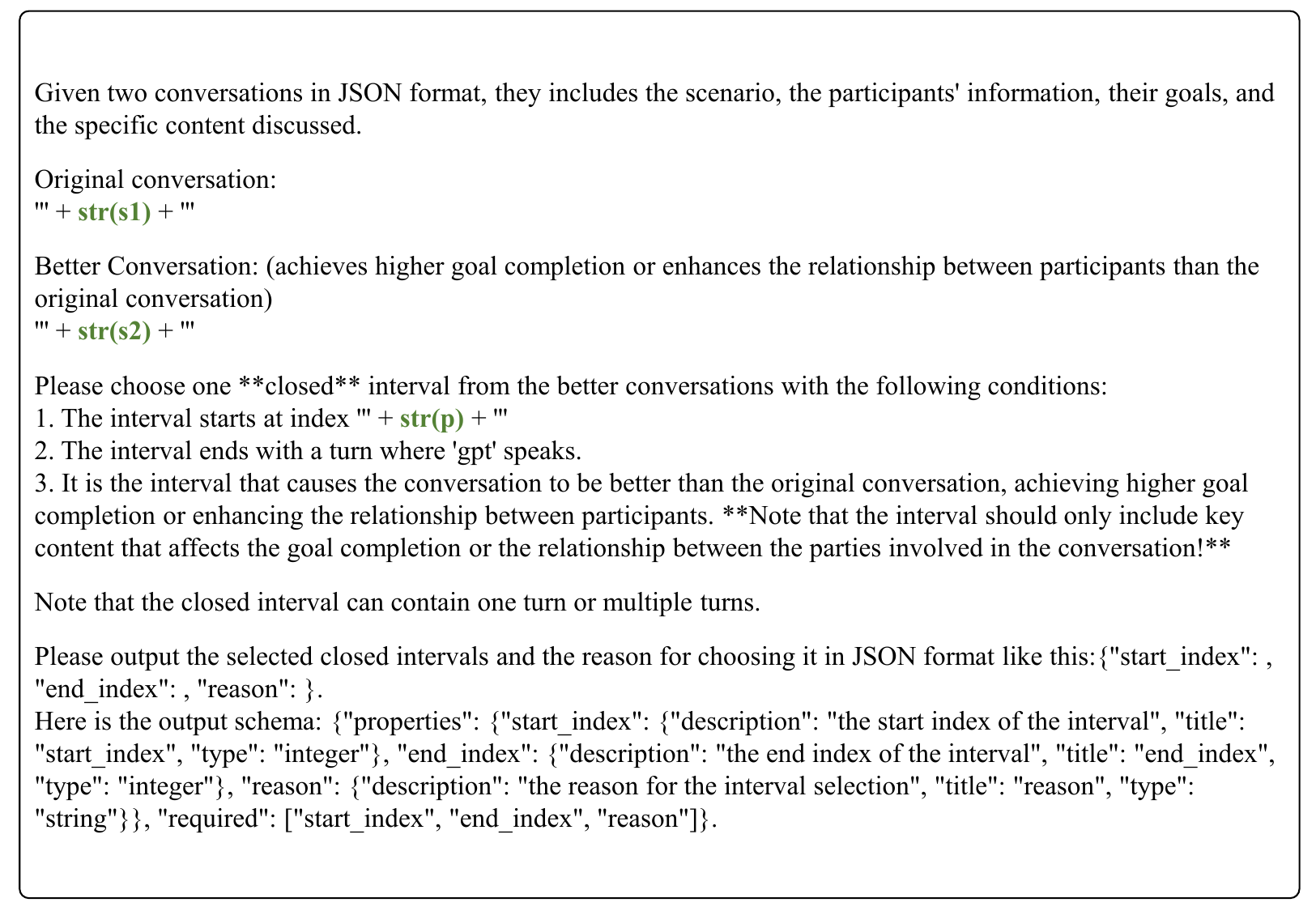}
  \caption{The prompt for segment selection with GPT-4o. $s_{1}$ represents the complete positive session, $s_{2}$ denotes the negative session, and $p$ indicates the error turn index, specifically the first differing turn between the positive and negative sessions.}
  \label{fg: location_prompt}
\end{figure*}

The prompts for error localization and segment selection with GPT-4o are presented in Figures \ref{fg: error_prompt} and \ref{fg: location_prompt}.

\section{Supplementary Experimental Setup}

\subsection{Versions of OpenAI LLMs}
\label{sec: version}
The OpenAI LLMs we used are as follows: GPT-4o-2024-08-06, GPT-4-turbo-2024-04-09, GPT-4o-mini-2024-07-18, and GPT-3.5-turbo-0125.

\subsection{Mistral Training Details}
\label{sec: mistral}
Consistent with the experimental settings of Llama, the maximum token limit is set to 4096, and AdamW optimization is employed for all training processes. During the SFT phase, the batch size is 32, the dropout rate is 0.2, and the learning rate is $3e^{-6}$ with $5\%$ warm-up ratio and a cosine decay schedule. For the training phase of SDPO, the batch size remains 32, $\beta$ in SDPO loss is 0.1, and the learning rate is $5e^{-7}$ with no warm-up but a cosine decay schedule. The construction of SDPO data for Mistral follows the same process as that for Llama.

\section{Additional Empirical Results}

\subsection{SOTOPIA Hard Subset}
\label{sec: hard}
SOTOPIA designates the more challenging portion of the dataset as the Hard subset, with detailed results presented in Table \ref{tb: hard}. The ranking of various methods on the Hard subset is generally consistent with their performance on the full dataset. SDPO still achieves the best results, indicating that SDPO enhances the agent's social intelligence across scenarios with different difficulty.

\begin{table*}
\centering
\begin{tabular}{lccccccc} 
\toprule
\multirow{2}{*}{Model}    & \multicolumn{2}{c}{Self-Chat} & \multicolumn{2}{c}{GPT-4o} & \multicolumn{2}{c}{GPT-4o-mini} & \multirow{2}{*}{AVG}  \\
\cmidrule(r){2-3}%
\cmidrule(r){4-5}%
\cmidrule(r){6-7}%
                          & GOAL         & REL            & GOAL          & REL             & GOAL         & REL                   &                       \\ 
\midrule
GPT-4-turbo               & 6.20          & 2.36           & 6.23          & 2.41            & 4.96         & 1.76                  & 3.99                  \\
GPT-4o                    & 6.10          & 2.14           & 6.10           & 2.14            & 5.15         & 1.59                  & 3.87                  \\
GPT-4o-mini               & 4.53         & 1.13           & 5.32          & 1.60             & 4.53         & 1.13                  & 3.04                  \\
GPT-3.5-turbo             & 3.52         & 0.65           & 4.54          & 1.26            & 3.65         & 0.89                  & 2.42                  \\
Llama-8B              & 4.94         & 0.33           & 6.17          & 1.65            & 5.03            & 1.26                     & 3.23                     \\
Llama-8B+BC           & 6.51         & 2.60            & 5.71          & 2.13            & 4.43         & 1.76                  & 3.86                  \\
Llama-3.1-8B+BC+DPO       & 6.69         & 3.00              & 6.37          & 2.43            & 5.18         & 1.68                  & 4.23                  \\
Llama-8B+BC+ETO       & 6.40          & 2.80            & 6.47          & 2.50             & 5.20          & 1.81                  & 4.20                  \\
Llama-8B+BC+DMPO      & 6.67         & 2.85           & 5.90           & 2.47            & 4.92         & \textbf{1.86}                  & 4.11                  \\
Llama-8B+BC+P-SFT     & 6.36         & 2.58           & 5.61          & 2.26            & 4.69         & 1.63                  & 3.86                  \\
Llama-8B+BC+SDPO      & \textbf{7.10} & \textbf{3.22}  & \textbf{6.69} & \textbf{2.78}   & \textbf{5.20} & 1.64\textbf{}         & \textbf{4.44} \textbf{}        \\ 

\bottomrule
\end{tabular}
\caption{The performance of various methods on SOTOPIA-Hard-Subset across the goal and relationship dimensions.}
\label{tb: hard}
\end{table*}

\begin{figure}[t]
\centering
\scalebox{0.69}{
\includegraphics[width=1\textwidth,trim=0.3cm 0cm 0cm 0cm, clip]{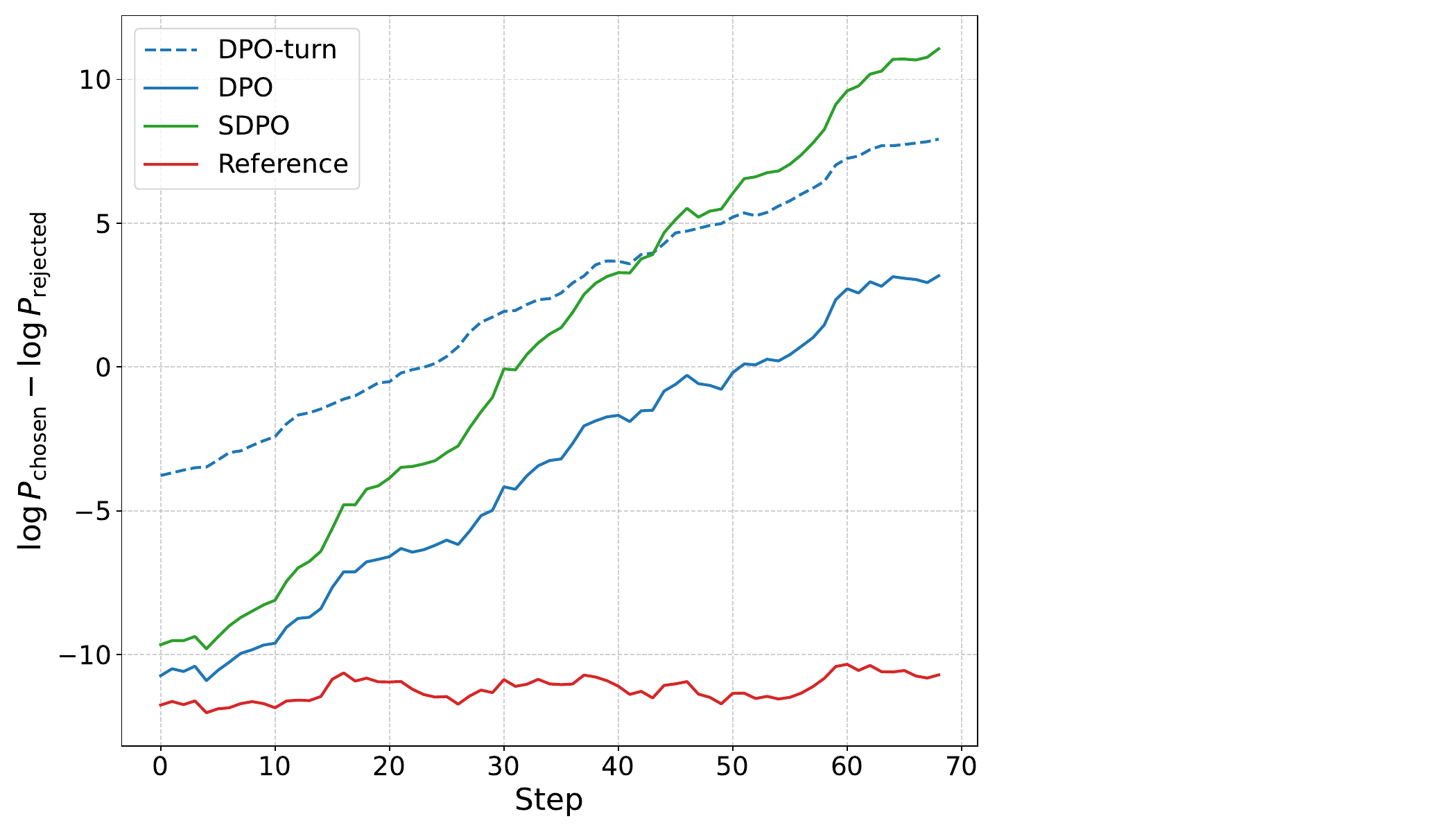}}
\caption{Variation in probability differences between positive and negative segments over training steps. DPO-turn represents the probability difference in the first turn in segments.}
\label{fg: necessity}
\end{figure}

\subsection{Necessity of Multi-turn Alignment}
\label{sec: proba}
After DPO adjusts the first-turn output probabilities for positive and negative segments, will the probabilities of positive segments increase and those of negative segments decrease in subsequent turns? To explore this, we plot the probability differences between positive and negative segments for DPO and SDPO during training, as shown in Figure \ref{fg: necessity} (only SDPO can be directly compared with DPO; therefore, ETO and DMPO are not mentioned here). The DPO-turn trajectory is nearly parallel to the DPO trajectory, indicating that DPO has almost no influence on the probability differences of subsequent turns. In contrast, the SDPO trajectory rises more steeply. These results demonstrate the necessity of explicitly modifying the probability distribution across turns within the entire segment, providing an explanation for the superiority of multi-turn alignment over DPO. 

\end{document}